\documentclass{ecai}
 
\usepackage{times}
\usepackage{soul}
\usepackage{url}
\usepackage[hidelinks]{hyperref}
\usepackage[utf8]{inputenc}
\usepackage[small]{caption}
\usepackage{graphicx}
\usepackage{amsmath}
\usepackage{amsthm}
\usepackage{booktabs}
\usepackage{algorithm}
\usepackage{algorithmic}
\usepackage[switch]{lineno}
\usepackage{adjustbox}
\usepackage{latexsym}

\begin{document}

\begin{frontmatter}
\title{Iterative Reward Shaping using Human Feedback for Correcting Reward Misspecification}
\author[1,2]{\fnms{Jasmina}~\snm{Gajcin}\thanks{Corresponding Author. Email: gajcinj@tcd.ie; Work done during internship at IBM Research Ireland}}
\author[1,2]{\fnms{James}~\snm{McCarthy}}
\author[1]{\fnms{Rahul}~\snm{Nair}}
\author[1]{\fnms{Radu}~\snm{Marinescu}}
\author[1]{\fnms{Elizabeth}~\snm{Daly}}
\author[2]{\fnms{Ivana}~\snm{Dusparic}}
\address[1]{IBM Research}
\address[2]{Trinity College Dublin}

\begin{abstract}
   A well-defined reward function is crucial for successful training of an reinforcement learning (RL) agent. However, defining a suitable reward function is a notoriously challenging task, especially in complex, multi-objective environments. Developers often have to resort to starting with an initial, potentially misspecified reward function, and iteratively adjusting its parameters, based on observed learned behavior. In this work, we aim to automate this process by proposing ITERS, an iterative reward shaping approach using human feedback for mitigating the effects of a misspecified reward function. Our approach allows the user to provide trajectory-level feedback on agent's behavior during training, which can be integrated as a reward shaping signal in the following training iteration. We also allow the user to provide explanations of their feedback, which are used to augment the feedback and reduce user effort and feedback frequency. We evaluate ITERS in three environments and show that it can successfully correct misspecified reward functions. 
\end{abstract}
\end{frontmatter}

\section{Introduction}
Reinforcement learning (RL) agents employ a trial-and-error learning process, with the aim to learn a policy maximizing the accumulated long-term reward. RL algorithms have shown great success in various fields \cite{arulkumaran2017}, and are being developed for use in real-life tasks, such as autonomous driving \cite{Kiran2022}.

A well-defined reward function is essential for successful training of an RL agent, as it directs the agent's learning. However, defining a suitable reward function is a notoriously difficult task. In multi-objective environments, it is often challenging to determine the correct weights for different, often conflicting desiderata that agent needs to optimize. A misspecified reward can lead to undesirable behavior or reward hacking, where the agent learns to ``game'' the system, such as a vacuum cleaner ejecting dust in a clean room, to maximize its cleaning time \cite{amodei2016concrete,pan2022effects}.  
While defining the appropriate reward function can be a challenging task, recognizing unwanted behavior stemming from a misspecified reward function is often more feasible. For example, while defining a reward function for an autonomous vehicle  is a challenging task, it is clear that a vehicle which increases the speed above acceptable limits is exhibiting undesired behavior.

In this work, we utilize human feedback to mitigate reward misspecification during training of RL agents. Our goal is to automate the iterative manual process that developers encounter in real-life tasks, where the initial, often misspecified reward is consecutively updated based on the observed behavior it produces. To that end, we implement ITERS (\underline{Ite}rative \underline{R}eward \underline{S}haping), an iterative approach which allows the user to observe agent's behavior at checkpoints during training, mark unwanted trajectories and provide explanations about why they assigned feedback to the selected trajectories. The marked trajectories and explanations are used to augment the feedback by generating a dataset of similar trajectories. Finally, a supervised learning model uses augmented trajectories to learn to recognize user-marked trajectories, and predict user's feedback for an input trajectory. The supervised learning model provides the reward shaping signal that is added to environment's reward in the following training iteration of the algorithm.

Previous work has shown that human feedback can improve convergence and sample complexity of RL algorithms under the assumption that it complements the environment's reward function \cite{knox2008tamer,guan2021widening,warnell2018deep,xiao2020fresh}. In our work, however, we explore the effect of human feedback when it directly contradicts the  environment's (misspecified) reward function. ITERS is thus applicable to scenarios where the user does not know the correct reward function and starts with a potentially misspecified reward. This corresponds to the process of devising a reward function for a real-life RL task where the correct reward function is not known beforehand and needs to be designed. Additionally, previous work on human-in-the-loop RL focuses on state-based feedback, assuming that the user knows a correct action in each state \cite{xiao2020fresh,knox2008tamer,warnell2018deep}. However, in complex environments with continuous action space, there might not be a large difference between performing two similar actions in a state, and behavior can only be evaluated on a more global level. For example, while it is difficult to evaluate a small change in the speed of an autonomous vehicle, it is more straightforward to recognize that increasing speed over multiple time steps leading to dangerous driving is undesirable behavior. To be able to capture more complex behavior we allow users to provide trajectory-based feedback. Moreover, while using explanations to augment feedback has been explored on the state level \cite{guan2021widening}, we allow for trajectory-level explanations which can include state features, actions and rule-based explanations.

Our contributions are as follows:

\begin{enumerate}
    \item We propose ITERS, an iterative reward shaping method for mitigating reward misspecification using human feedback.
    \item We provide an approach for augmenting trajectory-level user feedback and incorporating it with existing environment's reward function.
\end{enumerate}

Finally, we evaluate our work in three environments (including both single- and multi-objective environments, and discrete and continuous state spaces), where we show that ITERS can successfully correct the effects of a misspecified reward function. The implementation, evaluation scenarios and results of ITERS are available at: 
\href{https://github.com/anonymous902109/iters}{https://github.com/anonymous902109/iters}.

\section{Related Work}
Previous work has recognised the challenge of reward design and explored various ways of tackling it. Imitation learning attempts to learn a policy from expert demonstration through supervised learning \cite{zheng_imitation_2022}, whilst inverse reinforcement learning (IRL) instead proposes directly learning the reward function from expert demonstrations, and has been used to infer the reward function from human demonstration \cite{arora2021survey}. In similar vain, recent work by Qian et al., \cite{qian_learningrewards_2023} has explored iteratively learning a reward function using global performance metrics as a guide, in settings where reward specification is challenging, but global performance metrics are available. In contrast to such approaches, we propose to bring users into the loop to iteratively correct misspecified rewards through providing feedback, rather than from existing demonstrations or performance metrics.

Human-in-the-loop RL algorithms have utilized human feedback to learn a reward function \cite{knox2008tamer,warnell2018deep,xiao2020fresh,macglashan2017interactive}. In human-in-the-loop approaches, the user is asked to provide binary feedback based on whether agent chose the best action in a particular state. However, these approaches rely on the assumption that the user knows the best action in each state. In complex, high-dimensional, or continuous environments, it can be difficult for users to evaluate agent's behavior on such low level. Additionally, providing state-based feedback can be time-consuming and laborious for the user. While previous work allows trajectory-based feedback \cite{efroni2021reinforcement}, the approach requires the sum of rewards along the trajectory to be provided, which can be challenging for non-expert users to estimate. In  contrast, we only require binary trajectory-based feedback. 

Preference-based RL recognizes that requiring state-based feedback can be expensive and exhausting for the user \cite{christiano2017deep}. In preference-based RL, the user is periodically presented with a pair of trajectories, and asked to choose the one they prefer. This feedback is then used to learn a reward function corresponding to user's preferences. However, the approach has shown to have difficulty achieving good state coverage, and reduces human feedback to only binary signal \cite{ibarz2018reward}. In contrast, our approach allows the user to provide explanations, thus increasing the information gained from the feedback.  

Minimizing user engagement and effort is one of the main objectives for approaches relying on human input. EXPAND \cite{guan2021widening} aims to reduce the feedback frequency by allowing the user to provide explanations for state-based feedback in the form of important features. The explanation is used to augment the feedback by perturbing unimportant features, generating samples similar to
the original state. In this work, we extend this idea to trajectory-based feedback, and expand the possible explanations to include feature-based, action-based or rule-based explanations.

Most previous approaches that utilize human input use it instead of or as an extension of the environment's reward function. This means that the user's and environment's reward functions are the same and human feedback is used to amplify environment's reward and achieve faster convergence. However, in this work we explore the scenario where environment's reward is misspecified and the user's feedback contradicts it. Similar to our work, \cite{mindermann2018active} propose active inverse reward design, that aims to correct misspecified reward by iteratively asking the user to choose between two reward functions, to infer their preference. However, their approach assumes all reward components are known beforehand and that the user has a full understanding of their effects on the agent's behavior. In contrast, we do not assume expert knowledge of reward components and ITERS is applicable even if the misspecified reward omits important objectives.  

\section{ITERS: Iterative Reward Shaping}

In this section, we describe ITERS, an iterative reward shaping approach which utilises human trajectory-level feedback. At the start of the process, we assume a given, potentially misspecified, reward function $R_{env}$.  We also assume a user that can recognize wanted and unwanted behavior in the environment and can provide explanations for their feedback. This can either be a developer with the goal of discovering a suitable reward function for a specific task, or a user of the system wanting to adjust its behavior to their preferences. Through the iterations, ITERS uses human feedback to shape $R_{env}$.

At the start of every iteration, an RL agent is trained for a set number of steps $k$, to obtain a partially trained agent. Any RL algorithm can be used in this process, and we decide on DQN for all experiments. After partially training the agent for $k$ steps, a summary of a number of its best performing trajectories, according to the current reward, is presented to the user. The user can mark trajectories that exhibit unwanted behavior, as well as offer explanations for the assigned feedback. Explanations indicate which parts of the trajectory are important to the user, and are used to augment the marked trajectory and obtain an augmented dataset of similar trajectories. The augmented dataset is then provided as an input to the \textit{reward shaping model} $R_s$ -- a supervised learning model, which learns to generalize to the user's feedback across unmarked trajectories and predict the user's feedback based on the provided trajectory. Finally, the agent's training is resumed for another $k$ steps, and the environment's reward is shaped using the learned human feedback on a trajectory level. The iterative process is repeated until the user is satisfied with the agent's behavior, or the maximum number of iterations $n$ is reached. 

In the remainder of this section we first explain how the user's feedback is gathered (Section \ref{section3.1}). In Section \ref{section3.2} we explain the process of augmenting user feedback. Additionally, we show how reward shaping model is learned using the augmented dataset (Section \ref{section3.3}). Finally, we demonstrate how reward shaping signal is integrated in the following training iteration (Section \ref{section3.4}).

\begin{figure*}[ht]
    \centering
    \includegraphics[width=0.7\textwidth,height=9.0cm]{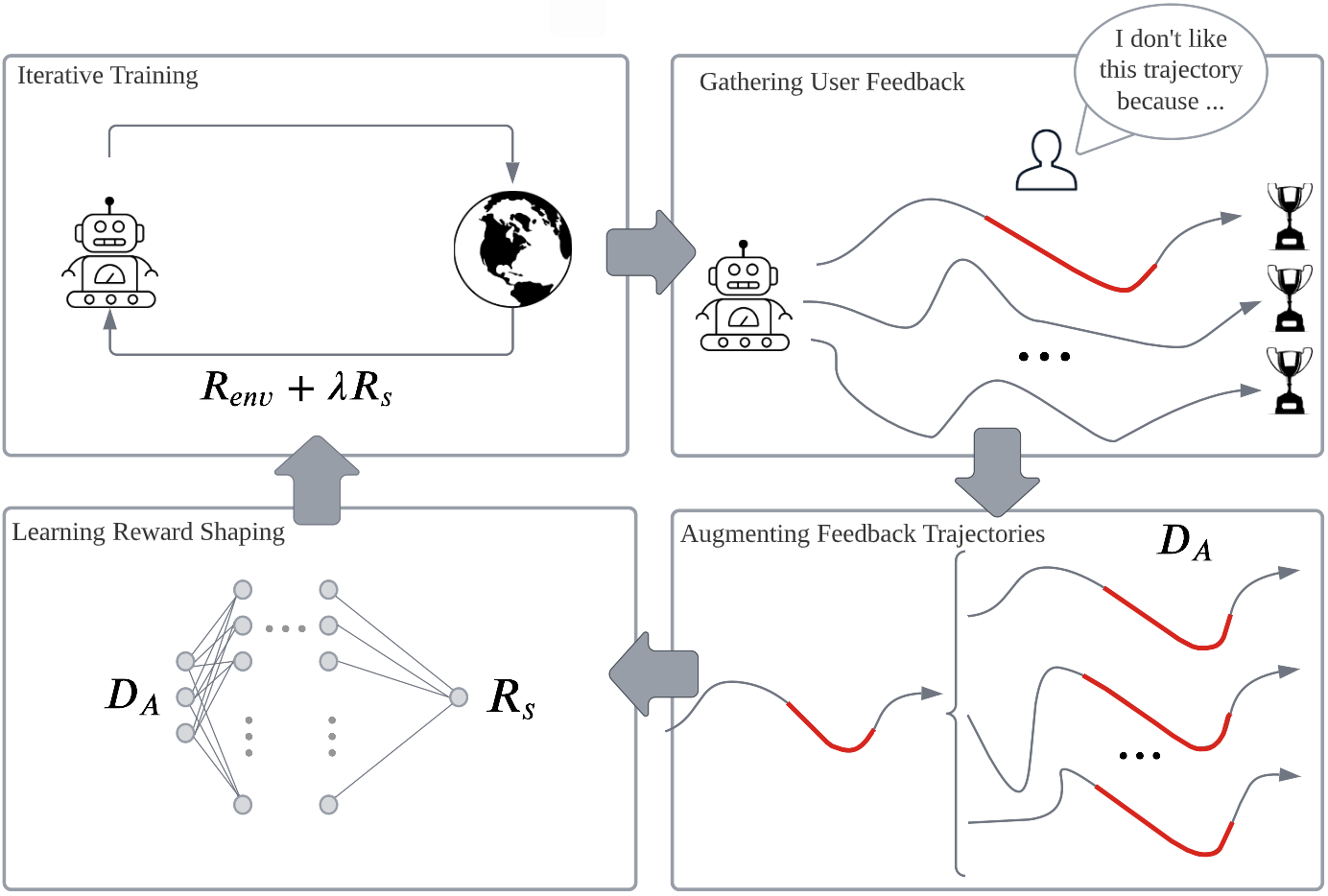}
    \caption{ITERS approach: The agent is iteratively trained, and user is presented with summary of agent's behavior at checkpoints. User can mark trajectories that reflect undesirable behavior and give explanations about their feedback. The explanations are used to augment marked trajectories by randomizing parts of the trajectory which are not important to the user. Augmented dataset $D_A$ is used to learn a reward shaping model $R_s$, which provides reward shaping signal in the following training iteration.}
    \label{ITERS}
\end{figure*}

The overview of the ITERS approach is shown in Figure \ref{ITERS}.

\subsection{Gathering User Feedback}
\label{section3.1}

At the start of each iteration, the agent is trained for a set number of steps $k$. In the first iteration, the agent is trained only on the environment's reward $R_{env}$, as no human feedback has been yet provided. In the following iterations, the agent will be trained on a shaped reward, as described in detail in Section \ref{section3.3}.

After partially training the agent for $k$ steps in each iteration, a summary of its behavior is presented to the user. As the agent is partially trained and its behavior likely contains some randomness, it is important to selectively summarize its behavior to the user. In this work, we unroll the agent's policy in the environment and gather the top performing $m$ trajectories. As a simplifying assumption, we focus on episodic environments, where each trajectory represents one episode. In Section \ref{future} we discuss the limitations of this assumption and potential summarization approaches that can capture agent's behavior in non-episodic tasks.  These trajectories demonstrate the behavior considered successful under agent's current reward function. 

The user can then provide feedback on a part of the trajectory where the agent exhibits unwanted behavior. Specifically, the user can mark the starting and ending step that bound the undesired behavior. We limit the size of the marked trajectories to contain at most $l$ time steps. As marked trajectories are used as an input to a supervised learning model $R_s$ as described in more detail in Section \ref{section3.3} we first ensure all trajectories are of the same length $l$. To that end, longer trajectories are clipped and shorter ones are padded with randomly generated states and actions. Additionally, we record the actual length of the marked trajectory and append it to the padded or shortened trajectory.

Additionally, the user can include an explanation, justifying why the specific trajectory has been reviewed. ITERS supports $3$ types of explanations:

\begin{enumerate}
    \item \textit{Feature-based}: The user can mark which state feature is important for the trajectory feedback. For example, if an autonomous vehicle unsafely increases its speed along a trajectory, the user can mark a speed-related feature as the reason for providing negative feedback.
    \item \textit{Action-based}: The user can provide feedback based on the actions along the trajectory. For example, if an autonomous vehicle changes lanes too frequently, user can mark a trajectory with too many lane changes according to their preference. To provide action-based explanation the user can mark the steering action which changes lanes along the trajectory. 
    \item \textit{Rule-based explanations}: The user can provide more complex explanations, based on state features or actions by employing simple SQL-like rules. For example, if the user considers frequent speed changes in an autonomous vehicle to be undesirable, they can provide a feedback along a trajectory $\mathcal{T}$ that features a number of speed changes and a rule-based explanation $E(\mathcal{T})$ in the form of an SQL-like rule:
    \begin{equation}
        E(\mathcal{T}): \textrm{COUNT}[v(s_t) \ne v(s_{t-1})]_{\mathcal{T}} > c_{max}
    \end{equation}
    
    where $c_{max}$ is the maximum allowed number of speed changes along the trajectory according to the user's preferences, and $v(s)$ is the agent's speed in state $s$.
\end{enumerate}

\subsection{Augmenting User Feedback}
\label{section3.2}

The feedback and explanations are used to augment marked trajectories. For each marked trajectory $\mathcal{T}$, unimportant elements are randomized, and values of important ones are kept constant, to obtain a dataset of augmented trajectories $\mathcal{A}(\mathcal{T})$, similar to the original one. For each marked trajectory $\mathcal{T}$, we generate an augmented dataset $\mathcal{A(\mathcal{T})}$ with $p$ trajectories.

The important elements of a marked trajectory are determined based on the user's explanations. For feature-based explanations, the important feature along $\mathcal{T}$ is rendered immutable, and remaining features are randomized to obtain $\mathcal{A}(\mathcal{T})$. For action-based explanations, the actions along the trajectory are kept constant. For rule-based explanation $E(\mathcal{T})$, $\mathcal{A}(\mathcal{T})$ is obtained by generating random trajectories that all satisfy the $E(\mathcal{T})$. If an important element is continuous, normal noise is added to it, to account for possible similar trajectories with slightly different important features. The actual length of a trajectory is always considered an important feature, and during augmentation is never randomized, as it indicates which part of the trajectory was marked by the user. The output from this stage  of ITERS in iteration $i$ is a dataset of augmented trajectories for each marked trajectory $\mathcal{T} \in \{\mathcal{T}^{(i)}_1, \cdots \mathcal{T}^{(i)}_m\}$:
\begin{equation}
    D^{(i)}_A = \mathcal{A}(\mathcal{T}^{(i)}_1) \cup \cdots \cup \mathcal{A}(\mathcal{T}^{(i)}_m)
\end{equation}

\begin{table*}[t]
    \centering
    \caption{Specification of the reward function $R_{env}$ and simulated user feedback for the GridWorld, highway and inventory environments.}
    \begin{adjustbox}{width=\textwidth}
    \begin{tabular}{ccccc} \toprule
         Task & Reward misspecification & User feedback & Explanation type & Explanation \\ \toprule
         GridWorld & Not penalizing turns & Trajectory with 4 consecutive turns & Action-based & 4 consecutive turn actions\\ \midrule
         Highway & Not penalizing lane changes & Trajectory where lane is changed more than two times in 5 steps & Feature-based & Agent's y coordinate \\ \midrule
         Inventory Management & Not including delivery cost & Trajectory where agent orders more than 5 times in a week & Rule-based & COUNT(ACTION $>$ 0) $>$ 5 \\ \bottomrule
    \end{tabular}
    \end{adjustbox}
    
    \label{expl}
\end{table*}

\subsection{Learning Reward Shaping Model}
\label{section3.3}

To generalize to the user's feedback for unseen trajectories, we train a neural network $R_s$ to predict the user's feedback based on the input trajectory. Specifically, the network predicts how many times the user marked the input trajectory. The neural network $R_s$ is updated in each iteration to include newly acquired feedback. Throughout iterations ITERS maintains a buffer $B^{(i)}$ of previously augmented trajectories and an array $M^{(i)}$ indicating in how many previous iterations trajectories in $B^{(i)}$ have been marked by the user. Intuitively, $M^{(i)}$ contains information on the strength of user's dissatisfaction with the marked trajectories. Initially, $B^{(0)}$ is populated with trajectories obtained by unrolling an agent $\mathcal{M}_{env}$, fully trained on the environment's reward function $R_{env}$. These initial trajectories provide a baseline for how agent would have acted if trained solely on the environment's reward $R_{env}$. Similarly, $M^{(0)}$ is populated with zeros, because human feedback has not yet been provided. Through iterations we update both $B^{(i)}$ and $M^{(i)}$ to include augmented dataset of user-marked trajectories. To update $R_s$ at iteration $i$, trajectory buffer $B^{(i)}$ is used as input and $M^{(i)}$ as the target.  

In each iteration $i$, we obtain a dataset of augmented trajectories $D^{(i)}_A$ from human feedback. We also generate an array $M_A$ which indicates how many times trajectories in $D^{(i)}_A$ have been marked by the user. If a user-marked trajectory $t$ can already be found among the previously marked trajectories $B^{(i-1)}$, we can check its value in $M^{(i-1)}$ and increment it in $M_A$. Otherwise, its value in $M_A$ will be $1$, indicating this is the first time this trajectory has been added. Similarly, we also need to update values in $M^{(i-1)}$ in case some previously marked trajectories are included again
in $D^{(i)}_A$. When checking if a marked trajectory can be found in the buffer, we only compare values of important features, as these are the only causes of the feedback according to the user and their explanations. Finally, we append the augmented dataset $D^{(i)}_A$ to the trajectory buffer and the marked array $M_A$ to $M^{(i-1)}$:

\begin{equation}
    \begin{aligned}
    B^{(i)} = B^{(i-1)} \cup D^{(i)}_A \\
    M^{(i)} = M^{(i-1)} \cup M_A
    \end{aligned}
\end{equation}

The reward shaping neural network $R_s$ is then trained on the updated buffer $B^{(i)}$ to predict $M^{(i)}$. Intiutively, $R_s$ predicts user's dissatisfaction with a certain trajectory, based on the how many times it has been marked as unwanted in the past. The details of the iterative training of $R_s$ are given in Algorithm \ref{Rs}.

\begin{algorithm}[t]
    \caption{Learning the reward shaping model $R_s$}
    \begin{algorithmic}
    \STATE \textbf{Input:} initial agent $\mathcal{M}_{env}$, number of iterations $n$
    \STATE \textbf{Output:} reward shaping model $R_s$
    
    \STATE $R_s = $ initNetwork() 
    \STATE $B^{(0)} = $ unroll($\mathcal{M}_{env}$)
    \STATE $M^{(0)} = [0, ..., 0] $
    
    \FOR{$i \in [1,\cdots, n]$}{
        \STATE $D^{(i)}_A =$ gatherAndAugmentFeedback()
        \STATE $M_A = [1, ..., 1]$ 
        \FOR{$t \in D^{(i)}_A$}{
            \STATE $m'_{ind} = $ findIndex($B^{(i-1)}, t$)
            \STATE $m'_t = M^{(i-1)}(m'_{ind})$ \STATE $M_A(t) = m'_t + 1$
        }\ENDFOR
        \FOR{$t \in B^{(i-1)}$}{
            \IF {$similar(t,D^{(i)}_A)$}{
            \STATE $M^{(i-1)}(t) = M^{(i-1)}(t) + 1$
            }\ENDIF
        }\ENDFOR
        \STATE $B^{(i)} = B^{(i-1)} \cup D^{(i)}_A$
        \STATE $M^{(i)} = M^{(i-1)} \cup M_A$
        \STATE $R_s$ = update($R_s, B^{(i)}, M^{(i)}$)
    }\ENDFOR
    \end{algorithmic}
    \label{Rs}
\end{algorithm}

\subsection{Integrating Reward Shaping Signal}
\label{section3.4}

The reward shaping model $R_s$ is trained to predict user feedback based on the input trajectory and is iteratively updated based on user's feedback. In each iteration, $R_s$ is then used to augment the environment's reward $R_{env}$. Specifically, during training, for each state $s_t$ and action $a_t$ the agent receives the reward:

\begin{equation}
    r_t = R_{env}(s_t, a_t) + \lambda R_s(\begin{bmatrix} (s_{t-l}, a_{t-l}) &  \cdots & (s_t, a_t)\end{bmatrix})
\end{equation}

Intuitively, the environment's reward is augmented by the trajectory-level reward corresponding to user feedback. The more times the user marks a specific behavior during iterations, the larger penalty will be received for that behavior in the future training. The strength of reward shaping depends on parameter $\lambda$. Larger values of $\lambda$ correspond to stronger influence of reward shaping, while for smaller $\lambda$ agent relies more on the environment's reward function. During the first training iteration output of $R_s$ is set to $0$, as no user feedback has been gathered at this point, and the agent is trained solely on the environment's initial reward $R_s$.

\section{Experiments}

We evaluate ITERS in three environments, where we explore whether human trajectory-level feedback can correct a misspecified reward function.

\subsection{GridWorld}
In a simple $5 \times 5$ GridWorld environment, an agent is tasked with reaching the goal. The agent can observe its coordinates, coordinates of the goal state, and its orientation. At each time step, the agent can choose between moving forward or changing its orientations for $90$ degrees clockwise. Each episode ends when the goal is reached, or the maximum number of steps is exceeded.

We initialize the environment with a misspecified reward function $R_{env}$ that gives +1 reward for reaching the goal, penalizes stepping forward with -1, but provides no negative reward for the turning actions (Table \ref{expl}). This reward function results in an agent which prefers to turn, never stepping forward nor reach the goal. In this environment our aim is to use human feedback to correct the misspecified reward function and obtain an agent which successfully navigates to the goal. We simulate human feedback by marking four-step trajectories in which the agent performs four turns and ends up in the same location and with the same orientation as in the start of the trajectory. In each iteration we mark at most one trajectory that fits this description. Additionally, we provide an action-based explanation for each marked trajectory, by marking the 4 consecutive turning actions that lead the agent to remain in the same position (Table \ref{expl}).

\subsection{Highway Environment}

The highway environment is a multi-objective environment with continuous state space \cite{highway-env}. The agent is tasked with driving safely on a 4-lane highway populated with other vehicles. The agent's goal is to drive as long as possible along the highway without crashing with the other vehicles. The 
continuous state captures the agent's presence, its $x$ and $y$ coordinates, and speed in $x$ and $y$ directions. Additionally, the agent has the same information about $4$ of its closest vehicles. Using $5$ discrete actions the agent can change lanes, increase or decrease speed or remain idle. 

The agent must balance between avoiding other vehicles and maintaining speed which is within the safe limits. The weighted sum of these two objectives makes the environments reward function $R_{env}$. We propose that this reward function is misspecified as it does not include a penalty for lane changes, and simulate a user that does not want more than two lane changes within a five-step trajectory. To simulate such feedback, in each iteration we mark every five-step trajectory where the lane has changed more than two times. For each marked trajectory we provide a feature-based explanation by marking the y coordinate feature of the agent, which changes when the lane is changed (Table \ref{expl}).

\subsection{Inventory Management}

In the inventory management environment an agent is tasked with maintaining and selling its stock by buying and selling items, depending on their demand. This is an environment with discrete state and action space. At each time step, the agent only has knowledge of its current stock, and can choose an action indicating how much stock should be bought. After buying the stock, the agent immediately sells all items that are in demand and can claim the profit. The demand is generated randomly using the Poisson  distribution with mean $30$. Each episode lasts for 14 timesteps, corresponding to two-week inventory management.

This is another multi-objective task, where the agent needs to balance between the cost of buying stock and the profit of selling items, while making sure to satisfy the demand. We initialize a reward function $R_{env}$, which in each step rewards agent for selling stock and incurs a cost for all items bought. $R_{env}$  also includes a penalty for when agent's stock is too low to satisfy the demand. From the perspective of some businesses, this reward function might be misspecified, as it does not include the cost of stock delivery. For that reason, we simulate a user who wants to limit the deliveries to a maximum of $5$ times within a week. We simulate the feedback of this user by marking all 7-day trajectories which feature more than 5 orders. With each marked trajectory we provide a rule-based explanation, which indicates a penalty should be applied when the number of actions larger than 0 is larger than 5 within a 7-day period (Table \ref{expl}).

\section{Evaluation}

In each environment, we start with a misspecified reward function $R_{env}$ (Table \ref{expl}) and use simulated task-specific human feedback to correct the agent's behavior. For all experiments we choose a DQN to represent agent's policy. However, ITERS is completely model-agnostic and any RL model could be used in its place. 

For evaluation purposes we assume access to the true reward function $R_{true}$ corresponding to user's preferred reward function for each environment. We use this reward function to train an expert DQN agent $M_{true}$ and track the progress of ITERS through iterations in relation to $M_{true}$. Similarly, we train an agent $M_{env}$ using only the environment's misspecified reward function $R_{env}$. ITERS does not rely on the true reward function at any point during training, and $R_{true}$ is only accessed during evaluation to verify that the our approach can indeed steer away from the misspecified reward $R_{env}$ and towards the preferred $R_{true}$. 

As ITERS is the first approach that allows humans feedback to contradict the environment's reward and uses it to correct a misspecified reward function, we cannot compare it to previous work, which assumes human feedback is compatible with the environment's reward function. Instead, in this section we evaluate its different components. Specifically, we focus on three evaluation questions:

\begin{enumerate}
    \item Can ITERS learn a policy that achieves performance comparable to a policy $M_{true}$ trained on a true reward function $R_{true}$?
    \item How does the parameter $\lambda$ influence the performance of a policy trained with ITERS?
    \item What is the user effort required for ITERS to achieve performance comparable to the policy $M_{true}$ trained on the true reward function $R_{true}$? 

\end{enumerate}

\begin{table}[t]
\caption{Parameters of ITERS algorithm used in GridWorld, highway and inventory management environments.}
    \centering
    \begin{adjustbox}{width=\linewidth}
        \begin{tabular}{cccc} \toprule
              & GridWorld &  Highway &  Inventory Management \\ \toprule
             Training time steps ($k$) & 20 000 & 10 000 & 10 000\\ \midrule
             Number of iterations ($n$) & 50 & 50 & 30\\ \midrule
             Number of summary trajectories ($m$) & 10 & 10 & 10\\ \midrule
             Marked trajectory length ($l$) & 5 & 5 & 7\\  \midrule
             Size of augmented dataset ($p$) & 10 000 & 10 000 & 10 000 \\ \midrule
         Augmentation 
 random normal noise mean ($\mu$) & - & 0 & -\\ \midrule
         Augmentation random normal noise std.  ($\sigma$) & - & 0.001 & -\\
             \bottomrule
        \end{tabular}
    \end{adjustbox}
    \label{params}
\end{table}
We explore questions $1$ and $2$ in Section \ref{5.1}, and question $3$ in Section \ref{5.2}. The parameters used by ITERS in different tasks are provided in Table \ref{params}.

\begin{figure*}[t]
    \centering
    \includegraphics[width=0.33\textwidth]{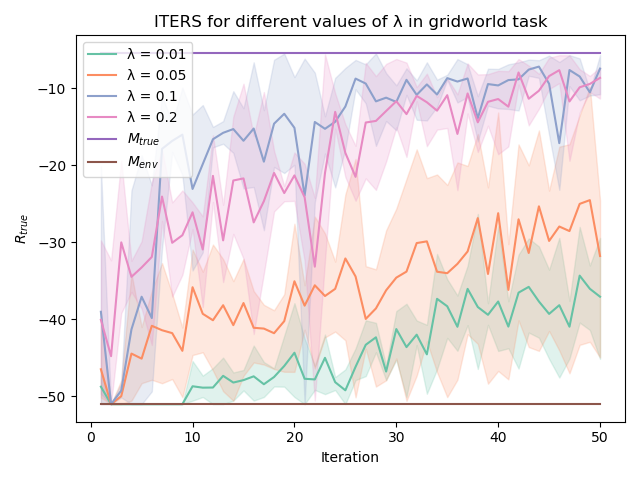}
    \includegraphics[width=0.33\textwidth]{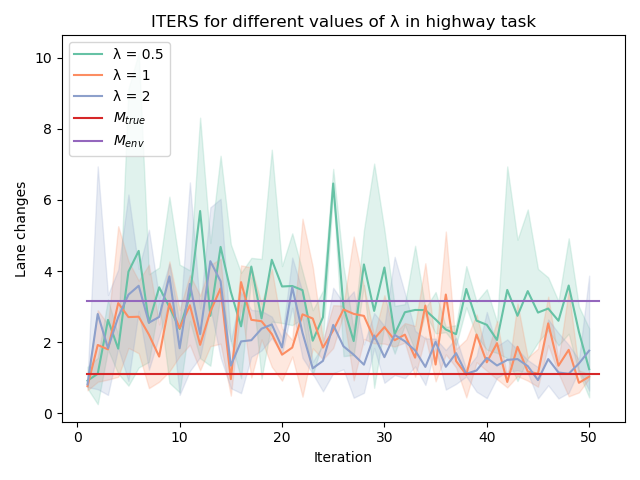}\includegraphics[width=0.33\textwidth]{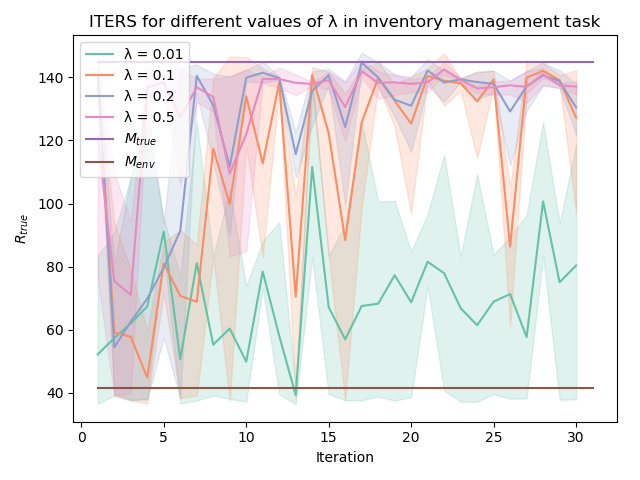}
    
    \caption{Convergence of ITERS algorithm to the behavior of $M_{true}$, trained on the true reward function $R_{true}$ for different values of parameter $\lambda$ in GridWorld, highway and inventory management tasks. For easier visualization of progress, the average number of lane changes within a 5-step trajectory is visualized for the highway task.}
    \label{convergence}
\end{figure*}

\subsection{Convergence to $M_{true}$ and the Choice of $\lambda$}
\label{5.1}

We start by evaluating the convergence of a policy trained with ITERS to the behavior of an expert policy $M_{true}$ trained on the true reward function $R_{true}$. For each task, we test ITERS with different values of the parameter $\lambda$ which determines the strength of human feedback. For each task and each value of $\lambda$ we run three experiments with different random seeds, and average the outcomes. ITERS starts with a misspecified reward function $R_{env}$ and runs for $n$ iterations (Table \ref{params}) using simulated task-specific human feedback to correct the effects of $R_{env}$. The results for all tasks are available in Figure \ref{convergence}. 

In the GridWorld environment, we run ITERS for $50$ iterations, with 4 different values of $\lambda \in \{0.01, 0.05, 0.1, 0.2\}$. The fastest reward shaping is achieved for $\lambda = 0.1$. After $50$ iterations, policy has digressed completely from environment's misspecified reward, and achieves performance comparable to the expert agent $M_{true}$. For lower values of $\lambda \in \{0.01, 0.05\}$, the effect of reward shaping is smaller, and the agent does not reach expert performance within $50$ iterations. Somewhat surprisingly, the largest value of $\lambda = 0.2$ does not deliver the fastest convergence. We believe that this is because  strong reward shaping signal can discourage optimal behavior as every action is met with a large penalty. 

In the inventory management environment, we run ITERS for $30$ iterations, and with four different values of $\lambda \in \{0.01, 0.1, 0.2, 0.5\}$. In this task, ITERS achieved the fastest convergence for the largest value of $\lambda=0.5$, where it reaches the performance of $M_{true}$ within $30$ iterations (Figure \ref{convergence}). For smaller values where $\lambda \in \{0.2, 0.1\}$ ITERS takes longer, but learns a policy with performance comparable to that of $M_{true}$ within $30$ iterations. The smallest value of $\lambda = 0.01$ results in a policy which achieves a better performance than $M_{env}$, but does not reach the performance of $M_{true}$.

For the highway environment we visualize the number of lane changes instead of $R_{true}$ (Figure \ref{convergence}) as there is a small difference in overall performance between models trained on the different reward functions, due to reward normalisation, making it difficult to visualize differences in their behavior. However, a clear difference in number of lane changes within a 5-step trajectory can be seen between   $M_{env}$ trained on $R_{env}$ and $M_{true}$ trained on $R_{true}$ (Figure \ref{convergence}). As $R_{env}$ does not include a penalty for lane changes, $M_{env}$ performs them more frequently than $M_{true}$. To correct the misspecification of the initial reward function $R_{env}$ we run ITERS for three values of $\lambda \in \{0.5, 1, 2\}$. For each value of $\lambda$ we run experiments with three different random seeds, each for $50$ iterations. Results are shown in Figure \ref{convergence}. ITERS shows the best results for  $\lambda = 2$ in the highway environment, where the agent trained by ITERS converges to the number of lane changes of an agent trained on the true reward function $M_{true}$. Smaller values of $\lambda \in \{0.5, 1\}$ however are not enough to reduce the number of lane changes within $50$ iterations.

In Appendix A of the supplementary material we explore examples of differing behaviour between $M_{env}$ and the agent trained by ITERS, in the highway environment. An interpretable Boolean rule-based model is trained on a dataset of state-action pairs to predict $M_{env}$'s next action, giving a global summary of the agent's policy in the form of Boolean rule sets. Agreement between the rules and each agent's selected action is investigated, showing differences between the two agent's policies.

\subsection{Feedback Frequency and User Effort}
\label{5.2}

\begin{table}[t]
\caption{The average number of marked trajectories provided to ITERS for GridWorld, highway and inventory management tasks.}
    \centering
    \begin{adjustbox}{width=0.7\linewidth}
        \begin{tabular}{ccc} \toprule
             Task & $\lambda$ & Average Feedback \\  \toprule
             GridWorld & 0.1 & 3 \\ \midrule
             Highway & 2 & 123.33\\ \midrule
             Inventory Management & 0.5 & 4.34 \\
             
             \bottomrule
        \end{tabular}
    \end{adjustbox}
    
    \label{feedback}
\end{table}

One of the greatest challenges of human-in-the-loop approaches is collecting a large amount of feedback without overwhelming the user. To evaluate the user effort needed for ITERS to achieve convergence to $M_{true}$, we record the number of simulated marked trajectories that ITERS requires in different environments. In Table \ref{feedback} we present the average number of  marked trajectories provided to the ITERS algorithm by the simulated user in the GridWorld, highway and inventory management tasks. Due to space constraints we present the average number of marked trajectories provided to ITERS only for the best performing value of learning parameter $\lambda$.

In the GridWorld environment, ITERS achieved best results with $\lambda = 0.1$. For this value of parameter $\lambda$ ITERS only required an average of $3$ marked trajectories over $50$ iterations (Table \ref{feedback}). This is a simple environment where the unwanted behavior is always exhibited in the same way through four consecutive turning actions. For this reason, ITERS can easily recognize unwanted trajectories, and repeated feedback is only needed to indicate that reward shaping should increase in strength in the next iterations.

In the inventory management environment, ITERS performs best for $\lambda = 0.5$. Similarly to GridWorld, unwanted behavior of ordering more than $5$ times within a 7-day period can be exhibited only in a handful of ways. For that reason, only an average of $4.34$ trajectories have been marked by the simulated user over $30$ iterations to achieve convergence to $M_{true}$ performance (Table \ref{feedback}).  

The best result in the highway environment is achieved for $\lambda=2$. The highway environment is the most complex of the examined environments,  and the unwanted behavior (Table \ref{expl}) can be exhibited in a large number of different ways. Additionally, this is an environment with a continuous state space, making it more difficult to generalize to all possible variations of the unwanted behavior. For this reason, highway environment requires the largest average number of user-marked trajectories for successful reward shaping. Specifically, in the highway environment, on average $123.33$ trajectories are provided to ITERS (with $\lambda = 2$) over $50$ iterations (Table \ref{feedback}).

\section{Conclusions and Future Work}
\label{future}
In this work, we proposed ITERS, an iterative approach for mitigating the effects of a misspecified reward using human feedback in RL. ITERS is the first human-in-the-loop approach that utilizes user feedback to correct reward misspecification, instead of augmenting the environment's reward. ITERS enables the user to provide trajectory-level feedback along with the explanations justifying their evaluation of agent's behavior and provides an approach for augmenting user feedback, in order to decrease user effort. We evaluated ITERS in three tasks, and found that it can correct the misspecified reward with only a handful of provided feedback trajectories. 

Our work, however, enables a limited number of specific explanation types to be provided, and uses simple randomization of unimportant features to augment the feedback. In future work we hope to extend the space of supported explanation types and generate more realistic augmented trajectories. Additionally, ITERS is limited to episodic environments where agent's behavior can be summarized in the form of episode trajectories. For ITERS to be applicable to non-episodic tasks alternative summarization methods that extract agent's behavior from continuous trajectories are needed. For example, user can be presented with the most interesting parts of agent's continuous behavior \cite{sequeira2020interestingness} in order to provide their feedback on the the agent's most unusual learned behaviors. Values for $\lambda$ were manually explored during training, in future work we hope to explore dynamically adjusting this hyperparameter during the training process. Finally, while we have shown that simulated human feedback can be used to correct a misspecified reward, we hope to explore the usefulness of ITERS in a user study. 

\ack This publication has emanated from research supported in
part by a grant from Science Foundation Ireland under Grant
number 18/CRT/6223 . For the purpose of Open Access, the
author has applied a CC BY public copyright licence to any
Author Accepted Manuscript version arising from this submission.

\bibliography{ecai23}

\end{document}